\documentclass{llncs}
\usepackage{epsf}
\usepackage{amsmath}
\usepackage{multirow}
\usepackage[linesnumbered]{algorithm2e}
\usepackage{url}
\usepackage{ulem}
\usepackage{fancybox}
\usepackage{authblk}
\newcommand{\superscript}[1]{\ensuremath{^{\textrm{#1}}}}
\def\wu{\superscript{1}} 
\def\wg{\superscript{2}}
\def\wp{\superscript{3}}
\def\comma{\superscript{,}}
\begin{document}




\title{Topics and Label Propagation:\\Best of Both Worlds for Weakly Supervised\\Text Classification}

\author{Sachin Pawar\wu\comma\wg, Nitin Ramrakhiyani\wu \\ Swapnil Hingmire\wu\comma\wp, Girish K. Palshikar\wu}
\affil{\tt \{sachin.pawar, nitin.ramrakhiyani\}@tcs.com}
\affil{\tt \{swapnil.hingmire, gk.palshikar@tcs.com\}@tcs.com}
\affil{\wu TCS Research, Tata Consultancy Services, Pune-13, India}
\affil{\wg Dept. of CSE, Indian Institute of Technology Bombay, Mumbai-400076, India}
\affil{\wp Dept. of CSE, Indian Institute of Technology Madras, Chennai-600036, India}
\institute{}

\maketitle

\begin{abstract}
We propose a Label Propagation based algorithm for weakly supervised text classification. We construct a graph where each document is represented by a node and edge weights represent similarities among the documents. Additionally, we discover underlying topics using Latent Dirichlet Allocation (LDA) and enrich the document graph by including the topics in the form of additional nodes. The edge weights between a topic and a text document represent level of ``affinity'' between them. Our approach does not require document level labelling, instead it expects manual labels only for topic nodes. This significantly minimizes the level of supervision needed as only a few topics are observed to be enough for achieving sufficiently high accuracy. The Label Propagation Algorithm is employed on this enriched graph to propagate labels among the nodes. Our approach combines the advantages of Label Propagation (through document-document similarities) and Topic Modelling (for minimal but smart supervision). We demonstrate the effectiveness of our approach on various datasets and compare with state-of-the-art weakly supervised text classification approaches.
\end{abstract}

\section{Introduction}
Text classification is an important area of Natural Language Processing (NLP) with applications ranging from automatic request routing to text understanding. It has also been one of the most active and competitive areas of research in NLP. In this work, we propose a novel weakly supervised method to solve document classification.

We use the Label Propagation algorithm~\cite{zhu2002learning} which works on an undirected graph and involves iterative propagation of labels from a few labelled nodes to large number of unlabelled nodes. The algorithm stops when label distributions at all nodes have converged. 

For Label Propagation, representation of documents in a graph and setting edge weights as similarity values among the documents is necessary. However, to achieve high accuracy, providing a good number of labelled documents is necessary. Document labelling can be an expensive and time-consuming activity and would require domain expertise. To do away with the cumbersome labelling of documents we propose to add topics learned over the documents in the graph and solicit labels only for topic nodes. Hingmire et al.~\cite{hingmire2013document} and Razavi et al.~\cite{razavi2013general} proposed labelling of topics instead of documents arguing that topic labelling invites lesser manual effort. We enrich the document similarity graph by adding labelled topics. We also introduce a topic influence parameter to control the topic enrichment process. Our algorithm LPA-TD (Label Propagation Algorithm - Topic Documents) constructs this topic enriched graph and runs Label Propagation on it to discover a classification of documents. The topic enriched graph is constructed for various configurations of the topic influence parameter and document similarities. Additionally, we experimented by constructing the topic enriched graph by dropping certain topic nodes which were incoherent and confusing to label. Closely seen, LPA-TD combines the power of topic modelling through smart manual tagging and iterative propagation of Label Propagation by harnessing document similarities.

We experiment on 4 public datasets from the 20Newsgroups (20NG) corpora and compare LPA-TD with multiple weakly supervised algorithms for text classification. We also compare LPA-TD with the performance of only Label Propagation (\textit{OnlyLPA}) using some labelled documents. LPA-TD outperforms the OnlyLPA baseline on all datasets and also outperforms the other algorithms on two out of four 20 NG datasets. We also perform experiments on a real-world dataset comprising of about 4000 grievances raised by employees of a large IT organization. The grievance text needs to be analysed by classifying it into four classes related to appraisals, compensation, finance and administration. Based on a manually created gold standard, LPA-TD performs at an encouraging macro-F1 of 78\% on this dataset.

The paper is organized as follows. In Section 2 we briefly describe the background of various techniques employed in the proposed LPA-TD algorithm. In Section 3, we describe the construction of the topic enriched graph and the topic influence parameter. Further in Section 4, we present details about the datasets, experimental setup, evaluation and analysis. Relevant related work is presented in Section 5. We finally detail some future work and conclude the paper.

\section{Background}
\subsection{Label Propagation Algorithm}
Zhu and Ghahramani~\cite{zhu2002learning} proposed the Label Propagation Algorithm which is a graph based semi-supervised method. It represents labelled and unlabelled instances as nodes in a graph with edges reflecting the similarity between nodes. The label information for any node is propagated to its nearby nodes through weighted edges iteratively and finally the labels of unlabelled examples are inferred when the propagation process is converged. The detailed version of the algorithm for transductive document classification is described in Algorithm~\ref{alg:lpa}.

\begin{algorithm}[h]
\KwData{1. $D_L$ (Set of labelled documents)\\2. $D_U$ (Set of unlabelled documents)\\3. $S$ ($n\times n$ Similarity matrix where $n=|D_L|+|D_U|$ and top $|D_L|$ rows correspond to labelled documents)\\4. $L=\{l_1,l_2,\cdots,l_m\}$ (Set of $m$ class labels)}
\KwResult{Label matrix $Y_{n \times m}$, where $Y_{ij}$ represents the probability of document $d_i$ having label $l_j$}
\caption{Label Propagation Algorithm for Transductive Document Classification}
\label{alg:lpa}
\tcc{Begin Initialization}
Define probability transition matrix $T$ such that $T_{ij} = \frac{s_{ij}}{\sum_k s_{kj}}$ which is the probability of jumping from $l_j$ to $l_i$\;
Define $\bar{T}$ as the row-normalized matrix of $T$ such that $\bar{T}_{ij}=\frac{T_{ij}}{\sum_k T_{ik}}$\;
Set iteration index $t=0$\;
Let $Y^0$ be the label matrix for $0^{th}$ iteration and $Y^0_L$ be its top $|D_L|$ rows and $Y^0_U$ be its remaining rows\;
Set $Y^0_{ij}=1$ if $d_i$ is labelled with $l_j$\;
Set values of $Y^0_U$ arbitrarily\;
\tcc{End Initialization}
Propagate the labels of any node to nearby nodes by $Y^{t+1}=\bar{T}Y^t$\;
Replace the values of top $|D_L|$ rows of $Y^{t+1}$ with $Y^0_L$\;
Set $t:=t+1$\;
Repeat steps 7 to 9 until $Y$ converges\;
\Return{$Y^{t}$}\;
\end{algorithm}

\subsection{Topic Modelling}
Topic modelling allows us to discover important and frequent themes or ``topics'' discussed in a large collection of text documents. The discovered topics provide an abstraction on the top of individual documents. Latent Dirichlet Allocation (LDA) \cite{journals/jmlr/BleiNJ03} is the simplest topic model. LDA and its variants have numerous applications in natural language processing, image processing, social network analysis etc. It is widely used to browse a large corpus of documents using the most probable words of each topic and the distribution over topics for each document \cite{DBLP:conf/icwsm/ChaneyB12}. LDA assumes following generative process for generating documents.

\begin{small}
\noindent 1. Select word probabilities ($\phi_t$) for each topic $t$:\\
\indent $\phi_t \sim \mathrm{Dirichlet}(\beta)$\\
2. Select topic proportions ($\theta_d$) for document $d$:\\
\indent $\theta_d \sim \mathrm{Dirichlet}(\alpha)$\\
3. Select the topic for each word position ($z_{d,n}$):\\
\indent $z_{d,n} \sim \mathrm{Multinomial}(\theta_d)$\\
4. Select the token for each word position ($w_{d,n}$):\\
\indent $w_{d,n} \sim \mathrm{Multinomial}(z_{d,n})$\\
\end{small} 
($\alpha$ and $\beta$ are Dirichlet priors for document-topics and topic-words distributions respectively.)\\

\subsection{Document Similarities}
In order to assign weights to edges connecting various documents, we need to devise a way of computing similarity between any two documents. Various document similarity measures and their effect on document clustering performance are discussed in detail by Huang~\cite{huang2008similarity}. For all our experiments, we have employed ``Cosine Similarity'' measure.

Let $D = \{d_1, d_2,\cdots,d_n\}$ be a set of $n$ documents and $W = \{w_1, w_2,\cdots,w_m\}$ be the set of $m$ distinct words (excluding stop-words and all words occurring in only one document), constituting the vocabulary of the corpus $D$. We represent each document $d_i$ by a vector $V_i$ of length $m$ whose $j^{th}$ component ($V_i[j]$) corresponds to the $j^{th}$ word in $W$ and is computed as,

\begin{equation*}
V_i[j] = TF(d_i,w_j)\cdot IDF(w_j)
\end{equation*}
where $TF(d_i,w_j)$ is number of times the word $w_j$ occurs in the document $d_i$ and $IDF(w_j)$ is computed using $ND(w_j)$, i.e. number of documents containing the word $w_j$ as $IDF(w_j)=\log\left(\frac{n}{ND(w_j)}\right)$.

For any two documents ($d_i$, $d_j$) and their corresponding vector representations ($V_i$, $V_j$), Cosine similarity between them is computed as follows:
\begin{equation*}
CosSim(d_i,d_j)=\frac{V_i\cdot V_j}{|V_i||V_j|}
\end{equation*}

\subsubsection{Document Graph Construction:} We construct a document graph where each node represents a document and an edge between any two documents indicates that the documents are similar. The degree of similarity between two documents is represented by assigning appropriate proportional edge weight. Higher edge weight indicates that there is a high similarity between the two documents. 

It was observed that each document generally has ``low'' cosine similarity with a large number of documents. In order to prevent label propagation among the dissimilar document nodes, we need to find some threshold on the document similarities such that if the similarity is below the threshold, then no edge will be added between such documents. We determine this threshold automatically for any set of documents. The threshold is determined such that at least 90\% documents within the set are connected to at least $K$ other documents. All the similarities below this threshold are forced to $0$ and hence no edge will be added in the document graph for a document pair with similarity below the threshold.

\section{LPA-TD: Proposed Approach for Text Classification}
In this section, we describe our novel approach for weakly supervised text classification.

\subsection{Weak Supervision by Labelling Topics}
The idea of manually obtaining labels for topics instead of instances was explored by Hingmire et al.~\cite{hingmire2013document,hingmire2014topic} and Razavi et al.~\cite{razavi2013general}. They observed that LDA topics are easily interpretable as they can be represented by their most probable words. A human annotator can provide the most suitable class label to each topic. The level of supervision in this case is quite low, as they found that a very few topics (typically twice the number of class labels) are generally enough. LDA topics uncover underlying semantic structure of the whole set of documents. Hence, even a few labelled topics, add significant information about most of the documents. Table~\ref{tab:topic_labels_example} shows the discovered topics and corresponding labels by assigned by a human annotator for the MEDICAL vs SPACE classification problem in 20 newsgroups dataset.

\begin{table}
\caption{Examples of topics discovered in the MEDICAL vs SPACE classification of 20 newsgroups dataset and corresponding labels assigned by a human annotator}
{\footnotesize
\begin{tabular}{p{11cm}c}
\hline
{\bf Topic (Most probable words)} & {\bf Label}\\
\hline
{\tt msg \underline{food} \underline{doctor} \underline{pain} day problem read evidence problems doesn \underline{blood} question case \underline{body} dyer} & Medical\vspace{0.5mm}\\
\hline
{\tt \underline{space} \underline{nasa} science program system data research information \underline{shuttle} technology \underline{station} center based sci theory} & Space\\
\hline
{\tt \underline{medical} \underline{health} water \underline{cancer} \underline{disease} number research information april care keyboard \underline{hiv} center reported \underline{aids}} & Medical\\
\hline
{\tt \underline{launch} \underline{earth} \underline{space} \underline{orbit} \underline{moon} \underline{lunar} \underline{nasa} high henry years \underline{spacecraft} long cost \underline{mars} pat} & Space\vspace{0.5mm}\\
\hline
\end{tabular}
}
\label{tab:topic_labels_example}
\end{table}

\subsection{Affinity between Topics and Documents:}
We use collapsed Gibbs sampling~\cite{Griffiths:Finding} to learn topics only using training documents. The collapsed Gibbs sampler for LDA gives topic-word distributions ($\phi_t$) for each topic and document-topic distribution ($\theta_d$) for each document.
We use $\theta_{d,t}$ i.e. probability of generating document $d$ by topic $t$ as the affinity between a training document $d$ in corpus and topic $t$. In other words, affinity measures the belongingness of a topic to a particular document. We use $\phi_t$ to infer $\theta_d$ for an unseen document using collapsed Gibbs sampling method proposed by Heinrich~\cite{heinrich2008parameter}. For a particular document, its affinities with all the topics are normalized so that they sum to 1.

\subsection{Topic-enriched Graph}

\begin{algorithm}[t]
\KwData{$D = \{d_1,d_2,\cdots,d_{N_D}\} = D_{train} \cup D_{test}$ (Set of $N_D$ documents containing training and test documents), $N_T$ (Number of topics to be discovered and later used for enrichment), $L=\{l_1,l_2,\cdots,l_m\}$ (Set of $m$ class labels), $\tau$ (Topic influence parameter)}
\KwResult{$D_L = \{<d_1,l_1>,<d_2,l_2>,\cdots,<d_n,l_n>\}$ where $l_1,l_2,\cdots,l_n \in L$}
$T:=$ Discover $N_T$ topics using LDA from documents in $D_{train}$\;
$T_L:= \{<t_1,l_1>,<t_2,l_2>,\cdots,<t_{N_T},l_{N_T}>\}$\tcc*[r]{Topic labels from human annotator}
$A_{dd}:=$ Compute document-document $N_D\times N_D$ similarity matrix\;
$A_{td}:=$ Compute topic-document $N_T\times N_D$ affinity matrix\;
\For{$i=1$ to $N_D$}{
$\mu:=0$\tcc*[r]{Total influence of the neighbouring documents}
\For{$j=1$ to $N_D$}{
$\mu:=\mu + A_{dd}[i][j]$\;
}
$c:=\frac{\tau\cdot\mu}{1-\tau}$\tcc*[r]{Multiplier for topic-document affinities so that for each document, fraction of influence by topic nodes is $\tau$}
\For{$j=1$ to $N_T$}{
$A_{td}[j][i]:= c\cdot A_{td}[j][i]$\;
}
}
\tcc{Similarity Matrix for Topic-enriched Graph}
$
A:=\left[
\begin{array}{cc}
A_{td} & 0 \\
A_{dd} & A^{T}_{td}
\end{array}\right]_{(N_T+N_D)\times (N_T+N_D)}
$\;
$Y_{(N_T+N_D)\times m} := LPA(T_L,D,A,L)$\;
$D_L := \Phi$\;
\For{$i=N_T+1$ to $N_T+N_D$}{
$j := $ Index of maximum probability in $Y[i]$\;
$D_L := D_L \cup \{<d_i,l_j>\}$
}
\Return{$D_L$}
\caption{LPA-TD: Label Propagation Algorithm on a Topic-enriched Document Graph}
\label{alg:lpa_td}
\end{algorithm}

We propose to enrich the document graph by adding a new node corresponding to each topic. As explained earlier, all these nodes are ``labelled'' nodes as the human supervision is provided at the topic level. All the other nodes representing documents are the ``unlabelled'' nodes. Each document node is connected to all topic nodes and the edge weight between a topic and a document is proportional to the ``affinity'' between them.

\subsubsection{Topic Influence Parameter:}
During the iterations of Label Propagation Algorithm, each document receives label distributions from the topic nodes as well as document nodes it is connected with. In other words, there are two sources of label information for a document node, i.e. ``labelled topics'' and ``similar documents''. To control the flow of label information from the two sources, we define a Topic Influence Parameter $\tau$. Through this parameter, the influence of topic information on a particular node can be fixed to be a specific fraction of the total edge weight incident on that node. Consider a document at which sum of incident edge weights from document nodes is $\mu$ (sum of incident edge weights from topic nodes is $1$ by definition as discussed in Section 3.2). To achieve the desired topic influence $\tau$, all the incident edge weights from topic nodes to the document are multiplied by a value $c$ changing the sum of incident edge weights from topic nodes to $c$. The value $c$ in turn can be expressed as a function of $\tau$ and $\mu$ as follows:
\begin{equation}
\tau = \frac{c}{\mu + c}\Rightarrow c = \frac{\tau * \mu}{1-\tau} 
\end{equation} 

Using a user-specified topic influence parameter, the topic-enriched document graph is constructed with appropriate edge weights. A classification of documents is then obtained by running Label Propagation Algorithm over this topic-enriched graph. Algorithm~\ref{alg:lpa_td} describes the LPA-TD algorithm in detail.

\section{Experimental Analysis}

\subsection{Datasets}
We report the performance of our experiments on corpora from the \textbf{20Newsgroups (20NG)} dataset. This dataset contains messages across twenty different UseNet discussion groups, posted over a period of time. These twenty newsgroups are grouped into 6 major clusters. We use the \textit{bydate} version of the 20Newsgroups dataset\footnote{\url{http://qwone.com/~jason/20Newsgroups/}}. This version of  the 20Newsgroups dataset contains 18,846 messages and it is sorted by the date of posting of the messages. The dataset is divided into training (60\%) and test (40\%) sets. We employ 4 different subsets of the 20NG dataset for our experiments, namely PC vs MAC, MEDICAL vs SPACE, POLITICS vs RELIGION and POLITICS vs SCIENCE. These subsets are fairly balanced in terms of representation of individual classes.

\subsection{Experimental setup}
We start by learning double the number of topics as number of classes over the training documents. Here, it is important to note that class information of training documents is not used. Training documents are used in unsupervised way only to learn the topics. Also, we do not use test documents for learning the topics to ensure fair evaluation. Test documents are only used for reporting results.

The learned topics are labelled by a single human annotator. The annotator is asked to label a topic with only one of the most appropriate class. 
We use the learned topics to compute a topic-document affinity matrix ($A_{td}$) for both the training and test documents.

Next, we construct the document to document similarity graph using two configurations: i) K = 1 and ii) K = 3 where K is minimum number of connected nodes for 90\% nodes in the graph as discussed in the Section 2.3. We include both the training and test documents in the document graph. To form the topic enriched graph we introduce the learned topics as additional nodes in the document similarity graph, with labels as assigned by the human annotator. We create edges from each document to all topics and experiment with multiple values of the topic influence parameter ($\tau$) for assigning edge weights. Results on the three best values of $\tau$ each for K = 1 and K = 3 are reported. 

As the process of learning topics is based on approximate inference, we carry out the topic learning, topic labelling and topic-document affinity computation processes 10 times. Hence, for a given configuration of the document similarity graph with a K and a $\tau$ value, the topic enriched graph is constructed 10 times. It is straightforward to see that the document-document similarity part remains same for all 10 runs, but topics and topic to document edges are added afresh for each run. We finally average the results over all 10 runs for a configuration and report them.

We compare the LPA-TD technique with multiple baselines presented below. In Table~\ref{tab:results}, we report the macro-F1 scores from the baselines and various configurations of LPA-TD. For the baselines requiring labelled documents, we provide them with the same number of labelled documents as number of topics to be labelled in LPA-TD.
\begin{itemize}
\item Expectation maximization with Naive Bayes for text classification proposed by Nigam et al.~\cite{Nigam:Text} with 4 randomly selected labelled documents
\item GE-FL~\cite{Druck:2008:LLF:1390334.1390436} with 10 labelled features, as reported by Hingmire et al.~\cite{hingmire2014topic}
\item TLC and ClassifyLDA, as proposed and reported by Hingmire et al.~\cite{hingmire2014topic}
\item Using only Label propagation: In this configuration, we only consider the document document similarity graph without topic enrichment and label as many number of documents as topics in LPA-TD. We then run Label Propagation on this graph and obtain the classification results for evaluation. We ensure that the most connected documents for both classes in the graph are labelled in equal proportion to ensure fairness and getting the best from this baseline.
\end{itemize}

\begin{table}[h]
\centering
\caption{Experimental Results}
\label{tab:results}
\begin{tabular}{p{3cm}p{1.2cm}p{1.8cm}p{1.8cm}p{1.8cm}p{1.8cm}}
\hline
                                                                            &     $\tau$ & pc-mac & med-space & politics-sci & politics-rel \\
\hline
TLC                                                                         &      & 0.68   & 0.943     & 0.911        & \textbf{0.922}        \\

ClassifyLDA                                                                &      & 0.641  & 0.926     & 0.899        & 0.892        \\

NB-EM                                                                       &      &    0.429    & \textbf{0.99}          & 0.474             & 0.466             \\
GE-FL                                                                       &      & 0.666  & 0.939     & 0.618        & 0.765             \\

OnlyLPA (K = 1)                &      &    0.486    &    0.919      &      0.539        &       0.559       \\
OnlyLPA (K = 3)                &      &    0.374    &    0.953       &     0.601         &      0.657        \\
\hline \hline
\multirow{3}{*}{\begin{tabular}[c]{@{}c@{}}LPA-TD\\   (K = 1)\end{tabular}} & 0.1  & 0.673  & 0.947     & 0.912        & 0.837        \\
                                                                            & 0.05 & 0.682  & 0.949     & 0.912        & 0.836        \\
                                                                            & 0.01 & 0.704  & 0.951     & 0.887        & 0.819        \\
\hline
\multirow{3}{*}{\begin{tabular}[c]{@{}c@{}}LPA-TD\\   (K = 3)\end{tabular}} & 0.2  & 0.661  & 0.948     & 0.918        & 0.840        \\
                                                                            & 0.1  & 0.673  & 0.950     & 0.916        & 0.839        \\
                                                                            & 0.05 & 0.671  & 0.949     & 0.903        & 0.823      \\
                                                                            \hline \hline
\multirow{3}{*}{\begin{tabular}[c]{@{}c@{}}LPA-TD-Coh\\   (K = 1)\end{tabular}} & 0.1  & 0.686  & 0.945     & 0.92        & 0.852        \\
                                                                            & 0.05 & 0.696  & 0.950     & 0.918        & 0.854        \\
                                                                            & 0.01 & \textbf{0.719}  & 0.954     & 0.887        & 0.849        \\
\hline
\multirow{3}{*}{\begin{tabular}[c]{@{}c@{}}LPA-TD-Coh\\   (K = 3)\end{tabular}} & 0.2  & 0.671  & 0.945     & 0.904        & 0.858        \\
                                                                            & 0.1  & 0.681  & 0.951     & \textbf{0.925}        & 0.860        \\
                                                                            & 0.05 & 0.674  & 0.953     & 0.918        & 0.853      \\
\hline
\end{tabular}
\end{table}

\subsection{Incoherent Topics}
As discussed earlier, to deal with approximate topic inference, we run all the experiments 10 times and report their average performance. However, for some particular runs, we observed that we get macro-F1 scores much below the average score for that configuration. Upon observing the learned topics in these runs, we found that some topics were incoherent and represent multiple classes. Hence, forcing them to one particular class label was introducing noise, in turn leading to poor performance. Table~\ref{tab:noisy_topics} shows examples of such incoherent topics.

\begin{table}
\caption{Examples of incoherent topics}
{\footnotesize
\begin{tabular}{p{10cm}c}
\hline
{\bf Noisy Topic (Most probable words)} & {\bf Dataset}\\
\hline
{\tt \dashuline{gun} \underline{space} \underline{nasa} \dashuline{president} \underline{launch} \dashuline{weapon} \dashuline{firearm} \underline{science} \dashuline{tax} system job \underline{earth} \underline{orbit} \dashuline{clinton} stephanopoulo} & \dashuline{politics}-\underline{sci}\\
\hline
{\tt drive disk hard system controller floppy rom \dashuline{bios} card port sound board power internal cable} & \dashuline{pc}-\underline{mac}\\
\hline
{\tt \dashuline{msg} \dashuline{food} idea high pat money long read remember thought \underline{moon} didn billion isn real} & \dashuline{med}-\underline{space}\\
\hline
{\tt system children objective \underline{moral} wrong \dashuline{fire} \dashuline{fbi} \underline{morality} \underline{koresh} opinions men doesn isn frank sex \dashuline{drugs}} & \dashuline{politics}-\underline{religion}\vspace{0.5mm}\\
\hline
\end{tabular}
}
\label{tab:noisy_topics}
\end{table}

In order to avoid adverse effect of incoherent topics on label propagation, we simply removed such topics from the topic-enriched graph while making sure that there is at least one topic mapped to each class. After removing incoherent topics from topic-enriched graph, we re-run LPA-TD algorithm. We refer this new approach as LPA-TD (Coherent) (\textit{LPA-TD-Coh}).

\subsection{Discussion}
Table~\ref{tab:results} shows experimental results of LPA-TD and LPA-TD (Coherent) along with other baselines. As we can observe from the results, LPA-TD outperforms both the configurations of the OnlyLPA baseline on all datasets. It also performs better than all other baselines on two datasets - PC vs MAC and POLITICS vs SCIENCE. 

PC vs MAC is considered to be the most difficult dataset from the 20NG corpora due to significant overlap of words seen in both classes. This overlap results from high semantic similarity between the classes. On this dataset, LPA-TD outperforms all the baselines comfortably, re-iterating the merit of the new technique.

It however, doesn't perform as well as the TLC and ClassifyLDA techniques in the POLITICS vs RELIGION dataset. A look at the topics learned in the 10 runs reveals mostly complex and fuzzy topics not attributable to a single class, which brings down the overall performance. Also, on the MEDICAL vs SPACE dataset, NB-EM outperforms LPA-TD but it performs quite poorly on other datasets. On the other hand, LPA-TD demonstrates a consistent performance across all the datasets.

\subsection{Case study on Employee Grievances}
We also carried out a case study to analyse a real-world industrial text dataset of grievances which were raised by employees of a major IT organization. The dataset contained about 4000 grievance descriptions related to areas like \texttt{finance}, \texttt{compensation}, \texttt{appraisals} and \texttt{administration}. However, no direct classification was available. So we got the dataset labelled from an HR executive for use as gold standard. Further, the grievances were sorted on the date of posting and we used first 70\% grievances for training and the rest for testing.
We tried out Naive Bayes and SVM classifiers using Weka\footnote{\url{http://www.cs.waikato.ac.nz/ml/weka/}} and obtained macro-F1 of about 80.9\% and 71.2\% respectively. 

Here, it is important to note that both Naive Bayes and SVM are supervised classifiers and required 70\% i.e. 2800 labelled grievances to achieve the above performance. Now, we employed our LPA-TD approach on this dataset. A few examples of the topics discovered and corresponding manual labels are shown in Table~\ref{tab:topic_labels_example_eg}. The topic enriched graph comprised of the 4000 grievance nodes along with the 8 learned topics. From the various configurations we tried, we obtained the best macro-F1 of 77.8\% for $K = 3$ and $\tau = 0.05$. This demonstrates a significant reduction in labelling effort (2800 grievances against only 8 topics) through use of the LPA-TD technique for comparable performance.

\begin{table}
\caption{Examples of topics discovered in the Employees Grievances dataset and corresponding labels assigned by a human annotator}
{\footnotesize
\begin{tabular}{p{10cm}c}
\hline
{\bf Topic (Most probable words)} & {\bf Label}\\
\hline
{\tt \underline{basic} \underline{salary} grade \underline{compensation} experience \underline{pay} higher variable \underline{allowance} months joined current designation letter mba} & Compensation\\
\hline
{\tt \underline{rating} project \underline{appraisal} \underline{performance} \underline{band} work \underline{appraiser} process team discussion client \underline{reviewer} final disagreement worked} & Appraisal\\
\hline
{\tt office \underline{bus} working \underline{admin} work \underline{food} \underline{cab} day provided \underline{facility} service card \underline{transport} \underline{canteen} issue} & Admin\vspace{0.5mm}\\
\hline
{\tt \underline{salary} amount month \underline{finance} \underline{claim} \underline{tax} account months received deducted paid ticket allowance days \underline{payroll}} & Finance\vspace{0.5mm}\\
\hline
\end{tabular}
}
\label{tab:topic_labels_example_eg}
\end{table}

\section{Related Work}
Previous work in semi-supervised text classification can be broadly categorized into 4 different types based on the way supervision is provided: i) Labelling a few documents, ii) providing a list of features that are highly indicative of each class label, iii) employing active learning and iv) labelling topics.

\subsection{Using Labelled and Unlabelled Documents}
In this method, labelled documents and a large number of unlabelled documents are used for learning the classifier. While estimating the parameters of the classifier certain assumptions about the distribution of labelled and unlabelled documents will have to hold.

\noindent \textbf{Cluster assumption} : if instances are in the same cluster, they are likely to be of the same class. In other words, if the data are generated by a mixture model following a generative process and a mixture component represents one or more classes then the instances generated by a mixture component are likely to have the same class labels. Due to unlabelled data, the mixture model contains both observed and hidden variables and its parameters are estimated by Expectation-Maximization (EM) algorithm~\cite{Dempster77maximumlikelihood}. Nigam et al.~\cite{Nigam:Text} used EM Algorithm for semi-supervised text classification with a naive Bayes classifier. 

\noindent \textbf{Low-density separation assumption}: the decision boundary of classification should lie in a low-density region. Using this assumption, Grandvalet and Bengio~\cite{DBLP:conf/nips/GrandvaletB04} proposed a maximum a posteriori (MAP) framework for learning a classifier using minimum entropy regularization. Another semi-supervised algorithm which makes this assumption for learning a text classifier using a small number of documents is \textit{Transductive Support Vector Machines (TSVMS)}\cite{Joachims:1999:TIT:645528.657646}.

\noindent \textbf{Manifold assumption:} the high-dimensional data lie (roughly) on a low-dimensional manifold. Graph based semi-supervised methods make the manifold assumption to construct a graph in which nodes are both the labelled and unlabelled instances and edge weights represent similarity between instances. The Label Propagation Algorithm~\cite{Zhu02learningfrom} discussed earlier falls in this category. Other graph based text classification approaches are by Subramanya and Bilmes~\cite{subramanya2008soft} and Wang and Zhang~\cite{wang2008label}.

\noindent \textbf{Multi-view assumption:} each instance has two or more ``different'' and ''independent'' views and each view is sufficient for good classification individually. Co-Training~\cite{Blum:1998:CLU:279943.279962} algorithm is based on this assumption. The Co-Training process initially constructs a weak classifier for each view using labelled instances, then each weak classifier is bootstrapped using unlabelled instances. 

\subsection{Incorporating Labelled Features}
Sometime it is easier for human annotators to describe a class of documents using a set of features than labelling large collections of documents. Liu et al.~\cite{Liu:2004:Text} proposed a text classification algorithm by labelling the most discriminative words. Eventually, these representative words are used to create a text classifier using the combination of naive Bayes classifier and the Expectation-Maximization (EM) algorithm. Other similar approaches are Schapire et al.~\cite{DBLP:conf/icml/SchapireRRG02} (based on AdaBoost), Wu and Srihari~\cite{Wu:2004:IPK:1014052.1014089} (generalization of SVM, Weighted Margin SVM) and Druck et al.~\cite{Druck:2008:LLF:1390334.1390436} (generalized expectation criteria based maximum entropy text classifier, i.e. GE-FL). 

\subsection{Using Active Learning}
Active learning~\cite{settles.tr09} systems attempt to overcome the labelling bottleneck by asking queries in the form of unlabelled instances to be labelled by a human annotator. Some important text classification approaches using active learning are Godbole et al.~\cite{DBLP:conf/pkdd/GodboleHSC04}, Raghavan et al.~\cite{Raghavan:2006:ALF:1248547.1248608} and Druck et al.~\cite{DBLP:conf/emnlp/DruckSM09}.

\subsection{Labelling Topics}
Hingmire et al.~\cite{hingmire2013document} proposed the idea of obtaining labels for topics instead of documents. They propose the ClassifyLDA algorithm where a topic model is leaned using LDA and one class label is assigned to each topic. They use the Dirichlet distribution to aggregate all the same class label topics into a single topic and automatically classify unlabelled documents based on their similarity with the aggregated topics. Hingmire and Chakraborti~\cite{hingmire2014topic} proposed the TLC algorithm which further improves the ClassifyLDA algorithm by allowing a topic to be labelled with multiple class labels instead of one.

\section{Conclusions and Future Work}
We proposed a weakly supervised text classification technique LPA-TD, based on Label Propagation and Topic Modelling. A topic enriched document graph is constructed for a set of documents where the only supervision is in the form of labelled topics. LPA-TD propagates labels over this topic enriched graph thereby exploiting benefits of both, document similarities and labelled topics.  We evaluated LPA-TD on 4 datasets of the 20NG corpora and compared with multiple baselines. LPA-TD outperforms all the baselines on 2 out of these 4 datasets, including PC vs MAC which is considered to be one of the most difficult for text classification. Compared to other baselines, LPA-TD demonstrates a consistent performance across all the datasets. We also showed that the issue of incoherent topics can be handled by removing them from the topic enriched graph, without degrading LPA-TD's performance. Furthermore, removal of such incoherent topics resulted in better performance.

In future, we plan to extend LPA-TD by allowing fuzzy class labels to topics which naturally represent multiple classes. This will ease the restriction of assigning only one class per topic. Additionally, we plan to devise topic quality measures for automatic detection of incoherent topics.

\bibliographystyle{splncs03}
\bibliography{ref}
\end{document}